\documentclass[12pt]{article}
\usepackage{setspace} 
\usepackage{amsmath}
\usepackage{amssymb}
\usepackage{graphicx}
\usepackage{enumerate}
\usepackage{natbib}
\usepackage{booktabs}

\usepackage{soul, xcolor} 
\usepackage{amsfonts}
\usepackage{amsthm}
\usepackage{bm}
\usepackage[labelformat=simple]{subcaption}
\usepackage{hyperref}

\newcommand{\norm}[1]{\lVert#1\rVert}

\DeclareMathOperator{\sgn}{sgn}

\newtheorem{property}{Property}
\newtheorem{theorem}{Theorem}
\newtheorem{definition}{Definition}

\title{A phase transition for finding needles in nonlinear haystacks with LASSO artificial neural networks}

\author{Xiaoyu Ma \\ Shandong University and University of Geneva\\
\texttt{maxiaoyu0416@gmail.com} \\
Sylvain Sardy\\ Department of Mathematics, University of Geneva\\ \texttt{sylvain.sardy@unige.ch}\\ 
Nick Hengartner \\ Los Alamos National Laboratory, Theoretical Biology and Biophysics group\\ \texttt{nickh@lanl.gov} \\
Nikolai Bobenko\\ Department of Mathematics, University of Geneva\\ \texttt{nikolai.bobenko@unige.ch}  \\
Yen Ting Lin\\ Los Alamos National Laboratory, Information Sciences group\\ \text{yentingl@lanl.gov}
}

\begin{document}

\maketitle

\begin{abstract}
To fit sparse linear associations, a LASSO sparsity inducing penalty with a single hyperparameter   provably allows to recover the important features (needles) with high probability in certain  regimes even if the sample size is smaller than the dimension of the input vector (haystack). More recently learners known as artificial neural networks (ANN) have shown great successes in many machine learning tasks, in particular fitting nonlinear associations. Small learning rate, stochastic gradient descent algorithm and large training set help to cope with the explosion in the number of parameters present in deep neural networks.
Yet few ANN learners have been developed and studied to find needles in nonlinear haystacks. Driven by a single hyperparameter, our ANN learner,  like for sparse linear associations,  exhibits a phase transition in the probability of retrieving the needles, which we do not observe with other ANN learners. To select our penalty parameter, we generalize the universal threshold of Donoho and Johnstone (1994) which is a better rule than the conservative (too many false detections) and expensive cross-validation. In the spirit of simulated annealing, we propose a warm-start sparsity inducing algorithm to solve the high-dimensional, non-convex and non-differentiable optimization problem. We perform precise Monte Carlo simulations to show the effectiveness of our approach.
   \end{abstract} 
   
 

\section{Introduction}


Over the past ten years, Artificial Neural Networks (ANNs) have
become the model of choice for machine learning tasks 
in many modern applications. Although not completely understood today, the beliefs of the reasons for their success are mathematical, statistical and computational. 

From the point-of-view of approximation theory, ANNs approximate well smooth functions. For instance a single hidden layer neural net with a diverging number of neurons is dense in the class of compactly supported continuous functions \citep{citeulike:3561150} and
the first error rate derived \citep{256500} motivates shallow learning (few layers) \citep{7069264,Kostadinov2018:EUVIP}.
Some results show that deep learning is superior to shallow learning in the sense that less parameters are needed to achieve the same level of accuracy for a smoothness and compositional class of functions, in which case deep learning avoids the curse of dimensionality; see \citet{Poggio2017} for a review. 
\citet{DBLP:journals/corr/abs-1901-02220} prove that
 deep neural networks provide information-theoretically optimal approximation of a very wide range of functions used in signal processing.
 \citet{Chen:1995:UAN:2325866.2328543} and related papers extend the results to wider classes of functions.
  Approximation bound of sparse neural network, that is with bounded network connectivity, has been studied for instance by \citet{DBLP:journals/corr/BolcskeiGKP17} who show a link between the degree of connectivity and the complexity of a function class.

In machine learning, the success of ANNs is huge and, in part,
can be attributed to their expressiveness or capacity (ability to fit a wide variety of functions).
The very large number of parameters and the layer structure of ANNs make them impossible to interpret.
ANNs are overparametrized with multiple distinct settings of the parameters leading to the same prediction.
So traditional measures of model complexity based on the number of
parameters do not apply.  This makes understanding and
interpreting the predictions challenging.
Yet in scientific applications, one often seeks to do just that.
In keeping with Occam's razor, among all the models with similar
predictive capability, the one with the smallest number of features
should be selected.  Statistically, models with fewer features 
not only are easier
to interpret but can produce predictors with good statistical properties
because such models disregard useless features that contribute only to higher variance.

Operationally, the model selection paradigm often 
uses a validation set or cross-validation (in which the data is randomly splitted, models are built on a training set and predictions are evaluated 
on the testing set).  While conceptually elegant, (cross-)validation sets
are of limited use if feature selection is of interest (it tends to select many irrelevant features), if fitting a single model is computationally expensive or if the sample size is small (in which case, splitting the data leaves few observations).  
ANNs and in particular deep ANNs are computationally expensive to fit, so cross-validation is an expensive way of selecting model complexity.
Aiming at good predictive performance on a test set, also known as \emph{generalization}, cross-validation is  a poor feature selector as it tends to select too many features.
In addition, quadratic prediction error from cross-validation 
 exhibits an unexpected behavior with models of increasing complexity:  as expected,
the training  error always decreases with increasing number of input features, but  
while the quadratic prediction error on the test set is at first U-shaped (initially decreasing thanks to  decreasing bias, and then increasing due to an excess of variance), it then unexpectedly decreases a second time.  This phenomenon known as \emph{double descent} has been empirically observed \citep{AdvaniSaxe2017,ClementHongler2019}.
For least squares estimation regularized by an $\ell_2$ ridge penalty \citep{ridgeHK}, double descent has been mathematically described for two-layer ANNs with random first-layer weights
by \citet{MeiMontanari2019} and \citet{HastieMRT2019}.
They show that for high signal-to-noise ratio (SNR) and large sample size, high complexity is optimal for the ridgeless limit estimator of the weights, leading to a smooth and more expressive interpolating learner. In other words, interpolation is good and leads to double descent, which after careful thinking should not be a surprise since the interpolating ANN becomes smoother with increasing number of layers, and therefore better interpolates between training data. Indeed with high SNR, the signal is almost noiseless, so a smooth interpolating function shall perform well for future prediction. But data are not always noiseless, and in noisy regimes, that is with low SNR and small sample size, \citet{MeiMontanari2019} observe that regularization is needed, as expected.

In this paper, we present an alternative to the use of a validation set geared towards identifying important features.
Specifically, we develop an automatic feature selection method for 
simultaneous feature extraction and generalization. For ease of
exposition,  we present
our novel method in the context of regression and classification, noting that the ideas can be ported beyond.
Our approach exploits ideas from statistical hypothesis testing that
directly focus on identifying significant features, and this 
without explicitly considering
minimizing the generalization error.  Similar ideas percolate the statistics
literature, see for example \citet{JS04}, \citet{CDS99}, \citet{Tibs:regr:1996} with LASSO, \citet{BuhlGeer11} who
propose methods for finding {\em needles in a haystack} in linear models. 
In this context, the optimized criterion is not the prediction error,
but is the ability  to 
retrieve the needles (i.e., relevant features).  Useful criteria include the stringent exact support recovery criterion, and softer criteria such as
the false discovery rate (FDR) and true positive rate (TPR).

Of course some regularization methods have already been developed to enforce sparsity to the weights of ANNs.   
For example, {\em dropout} leaves out a certain number of 
neurons to prevent overfitting, which incidentally can be used to perform 
feature selection \citep{DBLP:journals/corr/abs-1207-0580,DBLP:journals/jmlr/SrivastavaHKSS14}. 
Sparse neuron architectures can be achieved by other means:
\citet{pmlr-v70-mollaysa17a} enforce sparsity based on the Jacobian
and \citet{DeepFeatureSelection2016,10.55552976456.2976557,10.5555/2981562.2981711,MemoryBound2014,DBLP:journals/corr/abs-1901-01021} employ $\ell_1$-based LASSO penalty to induce sparsity.
 \citet{TSNNaT21} prune their ANNs based on a metric for neuron importance. \citet{Evci2019TheDO} discuss the difficulty of training sparse ANNs.
{\tt spinn} (sparse input neural networks) \citep{feng2019sparseinput} have a sparsity inducing penalty and is governed by two hyperparameters chosen on a validation set;
its improved version {\tt spinn-dropout}  (the former originally published in 2017) adds a dropout mechanism governed by an additional hyperparameter \citep{pmlr-v80-ye18b}. So  {\tt spinn-dropout} is a mix between $\ell_1$ and $\ell_0$ (subset selection) sparsity inducing method, similar to the pruning idea \citep{8578988,ChaoWXC20}. None of these learners have been studied in terms of phase transition in the probability of retrieving features.

All of these sparsity inducing methods suffer from two drawbacks: (1) the selection of the penalty parameter is rarely addressed, and when it is, the selection is based on a validation set, two methods  geared towards good generalization performance, not feature identification; (2) the ability to recover the ``right'' features has not been quantified through the prism of a phase transition in the probability of support recovery; only {\tt spinn} and {\tt spinn-dropout} consider criteria related to FDR and TPR.

This paper is organized as follows.  Section~\ref{sct:tf} presents the theoretical
framework  and defines our LASSO ANN learner.
Section~\ref{subsct:functionestimation} defines the statistical model and notation.
Section~\ref{subsct:SANN} reviews the LASSO sparsity paradigm for linear models and extends it to ANNs.
Section~\ref{subsct:activation} discusses the choice of activation functions.
Section~\ref{subsct:lambda} derives a selection rule for the penalty parameter, a  generalization of the universal threshold \citep{Dono94b}  to non-convex optimization due to the nonlinearity of ANN models.
Section~\ref{subsct:opti}  discusses optimization issues to solve the non-convex high-dimensional and non-differentiable optimization problem.
Section~\ref{sct:MCsimu} evaluates via simulations the ability of our method to exhibit a phase transition in the probability of exact support recovery for the regression task.
Section~\ref{sct:appli} evaluates with a large number of real data sets the ability of our method to perform feature selection and generalization for the classification task.
Section~\ref{sct:conclusion} summarizes the findings and points to future developments.

\section{LASSO ANN}  \label{sct:tf}

\subsection{Function estimation model and notation} \label{subsct:functionestimation}

Suppose $n$ pairs of ouput-input data $({\cal Y}, {\cal X})=\{({\bf y}_i,{\bf x}_i) \}_{i=1}^n$ are collected to learn about their association.
For example, in some medical applications (see Section~\ref{subsct:classif}), ${\bf x}\in{\mathbb R}^{p_1}$ is an input vector of $p_1$ gene expressions and ${\bf y}$ is any of $m$ cancer types that is coded as  a one-hot output vector of ${\mathbb R}^{m}$; classification aims at assigning the correct type of cancer given an input vector. In regression, $y$ is a scalar ($m=1$), for instance riboflavin production rate in a bacteria (see Section~\ref{subsct:regression}).

To model their stochastic nature, data can be modeled as realizations from the pair of random vectors $({\bf Y},{\bf X})$.
We assume the  real-valued response ${\bf Y} \in {\mathbb R}^{m}$ is 
related to  real-valued feature vector ${\bf X}\in  {\mathbb R}^{p_1}$
through the conditional expectation 
\begin{equation} \label{eq:condexpect}
{\mathbb E}[{\bf Y}\mid {\bf X}={\bf x}] = \mu({\bf x}),
\end{equation}
for some unknown function $\mu: {\mathbb R}^{p_1} \rightarrow \Gamma\subseteq {\mathbb R}^m$.
In regression,  $\Gamma={\mathbb R}$
and 
in classification,
$\Gamma=\{{\bf p}\in ({\mathbb R}^+)^m: \sum_{k=1}^m p_k=1\}$.

Many learners have been proposed to model the association $\mu$ between input and output. A recent approach that is attracting considerable attention models $\mu$ as a standard fully connected ANN with $l$ layers 
\begin{equation} \label{eq:muANN}
\mu_{\boldsymbol \theta}({\bf x})=  S_l \circ \ldots \circ S_1\left( {\bf x}\right), 
\end{equation}
where ${\boldsymbol \theta}$ are the parameters (see \eqref{eq:theta12}) indexing the ANN, and letting ${\bf u}={\bf x}$ at the first layer, the nonlinear functions $S_k({\bf u})=\sigma({\bf b}_k + W_k  {\bf u})$ maps the $p_k\times 1$ vector ${\bf u}$ into a $p_{k+1}\times 1$ latent vector obtained by applying 
 an activation function $\sigma$ component-wise, for each layer $k < l$. The vectors ${\bf b}_k$ are commonly named ``biases.'' 
The matrix of weights $W_k$ is $p_{k+1} \times p_k$ and the operation $+$ is the broadcasting operation. 

The last layer $k=l$ has two requirements.
First we must have $p_{l+1}=m$ to match the output dimension, so the last function is $S_l({\bf u})=G({\bf c}+W_l {\bf u})$ where $W_l$ is $m \times p_l$ and the intercept vector ${\bf c}\in {\mathbb R}^{m}$.
Second the function $G: {\mathbb R}^{m}\rightarrow \Gamma$ is a link function that maps ${\mathbb R}^{m}$ into the parameter space $\Gamma$. Commonly used link functions for classification are
\small
\begin{eqnarray}
  G({\bf u})&=&\left(\frac{\exp\{u_1\}}{\sum_{k=1}^{m}\exp\{u_k\}},
\cdots, \frac{\exp\{u_m\}}{\sum_{k=1}^{m}\exp\{u_k\}}\right)^{\rm T}  \label{eq:GSoft} \\
G({\bf u})&=&\left(\frac{\exp\{u_1\}}{\sum_{k=1}^{m-1}\exp\{u_k\}
+1},
\cdots, \frac{\exp\{u_{m-1}\}}{\sum_{k=1}^{m-1}\exp\{u_k\}+1},\frac{1}{\sum_{k=1}^{m-1}\exp\{u_k\}+1}\right)^{\rm T}\label{eq:Glogit}
\end{eqnarray}
\normalsize
respectively called Softmax and multiclass-Logit. For regression, $G(u)=u$.

The parameters indexing the neural network are therefore
\begin{equation} \label{eq:theta12}
 {\boldsymbol \theta}=(( W_1, {\bf b}_1, \ldots, {\bf b}_{l-1}), (W_2, \ldots, W_l,{\bf c}))=:({\boldsymbol \theta}_1, {\boldsymbol \theta}_2)
 \end{equation}
for a total of $\gamma=\sum_{k=1}^l p_{k+1}(p_k+1)$  parameters. The following property is straightforward to prove, but is crucial for our methodology; it is the reason for splitting ${\boldsymbol \theta}$ into ${\boldsymbol \theta}_1$ and ${\boldsymbol \theta}_2$.
\begin{property} 
 \label{prop:propconstant}
Assuming the activation function satisfies $\sigma(0)=0$, then setting  ${\boldsymbol \theta}_1={\bf 0}$ implies $\mu_{\boldsymbol  \theta}({\bf x})$ is the constant function $\mu({\bf x})={\bf c}$ for all~${\bf x} \in \mathbb{R}^{p_1}$. 
\end{property}

Our estimation goal for ${\boldsymbol \theta}$ is two-fold. First, we want to generalize well, that is, given a new input vector, we want to predict the output with precision. Second, we  believe that only a few features in the $p_1$-long input vector carry information to predict the output. So our second goal is to find needles in the haystack by selecting a subset of the $p_1$-long inputs. 
For many medical data treated in Section~\ref{sct:appli}, the input ${\bf x}$ is a vector of hundreds of gene expressions, and genetic aims to identify the ones having an effect on the output.  Feature selection has been extensively studied for linear associations, showing a phase transition between regimes where features can be retrieved with probability near one to regimes where the probability of retrieving the features is essentially zero. 
Our goal is to investigate such a  phase transition with ANN learners to retrieve features in nonlinear associations.

\subsection{Sparse estimation} \label{subsct:SANN}

Finding needles amounts to setting the weights to non-zero values when corresponding to  features in ${\bf x}$ that have predictive information.
So  we  seeks sparsity in the first layer on the weights~$W_1$.
For the other layers, large weights in a layer could compensate small weights in the next layer,  so 
we bound them by forcing unit $\ell_2$-norm; instead, \citet{feng2019sparseinput} and  \citet{pmlr-v80-ye18b} take the approach of a ridge penalty controlled by an additional hyperparameter fixed to the arbitrary value of $0.0001$. More precisely
we slightly modify the nonlinear terms in~\eqref{eq:muANN} and define  the $j^\text{th}$ nonlinear function $S_{k,j}$  in  layer $k$ as
\begin{equation} \label{eq:Skj}
S_{k,j}({\bf u})= \left \{ \begin{array}{ll}
\sigma\left({\bf b}_1^{(j)} + \langle {\bf w}_1^{(j)}, {\bf u} \rangle \right) & k=1\\
\sigma\left({\bf b}_k^{(j)} +\frac{ \langle {\bf w}_k^{(j)}, {\bf u}  \rangle}{\left \Vert{\bf w}_k^{(j)} \right\Vert_2} \right) & 1< k <l \\
G\left ({\bf c}+ \frac{ \langle {\bf w}_k^{(j)},{\bf u}  \rangle}{\left \Vert {\bf w}_k^{(j)}\right \Vert_2}\right ) & k=l  \end{array}
\right . , \quad j \in \{1,\ldots,p_{k+1}\},
\end{equation}
where ${\bf w}_k^{(j)}$ is the $j^\text{th}$ row of $W_k$.
At the last layer ($k=l$),  ${\bf c}$  plays the role of an intercept.

Sparsity in the first layer allows interpretability of the fitted model.
To enforce sparsity and  control overfitting, we take the conventional approach inspired by LASSO of minimizing a compromise between a measure~${\cal L}_n$ of closeness
to the data and a measure of sparsity $P$.
Owing to Property~\ref{prop:propconstant},
we estimate the parameters ${\boldsymbol \theta}=({\boldsymbol \theta}_1, {\boldsymbol \theta}_2)$ defined in~\ref{eq:theta12} by aiming the best local minimum
\begin{equation} \label{eq:L1}
\hat {\boldsymbol \theta}_\lambda = \arg \min_{ {\boldsymbol \theta}\in {\mathbb R}^\gamma} {\cal L}_n ({\cal Y} , {\mu}_{\boldsymbol \theta}( {\cal X})) + \lambda\ P({\boldsymbol \theta}_1)
\end{equation}
 found by a numerical scheme, where $\lambda>0$ is the regularization parameter of the procedure and $P$ is sparsity-inducing penalty \citep{10.1561/2200000015}. We stress out that our method is driven by the selection of a single regularization parameter $\lambda$, as opposed to other methods that use two or three hyperparameters \citep{pmlr-v80-ye18b,feng2019sparseinput}.
 
 Common loss functions between training responses ${\cal Y}$ and predicted values  ${\mu}_{\boldsymbol \theta}( {\cal X})$ include:  for $m$-class classification the cross-entropy loss
  ${\cal L}_n ({\cal Y} , {\mu}_{\boldsymbol \theta}( {\cal X}))=\sum_{i=1}^n {\bf y}_i^{\rm T}\log\mu_{\boldsymbol{\theta}}({\bf x}_i)$, where the $\log$ function is applied component-wise to the $m$-long vectors ${\bf y}_i$ and $\mu_{\boldsymbol{\theta}}({\bf x}_i)$; for regression  ${\cal L}_n ({\cal Y} , {\mu}_{\boldsymbol \theta}( {\cal X}))=\sum_{i=1}^n ( {y}_i- {\mu}_{\boldsymbol{\theta}}( {\bf x}_i))^2$.

 A commonly used penalty is the  $\ell_q$ sparsity-inducing penalty used by waveshrink~\citep{Dono94b} and LASSO~\citep{Tibs:regr:1996} for $q=1$ and group-LASSO~\citep{Yuan:Lin:mode:2006} for $q=2$
 \begin{equation} \label{eq:Pq}
 P({\boldsymbol \theta}_1)=
 \sum_{j=1}^{p_1}\|{\bf w}_{1,j}\|_q + \sum_{k=1}^{l-1} \|{\bf b}_k \|_q ,
  \end{equation}
 where ${\bf w}_{1,j}$ is the $j^\text{th}$ column of $W_1$.  
The choice $q=2$ forces the $j^\text{th}$ feature to be either on or off across all neurons. The former is more flexible since a feature can be on in one neuron and off in another one, so, in the sequel, we use $q=1$.
The reason for penalizing the biases as well is that the gradient of the loss function with respect to the biases at zero is zero and that the hessian is positive semi-definite (see Appendix~\ref{app:Hessian}), hence no guaranteeing a local minimum.

ANNs are flexible in the sense that they can fit nonlinear associations. A more rigid and older class of models that has been extensively studied  is the class of linear models
 \begin{equation} \label{eq:lm}
\mu_{\boldsymbol \theta}^{\rm lin}({\bf x})= c+\sum_{j=1}^{p_1} \beta_j x_j,
\end{equation} 
where here the set of parameters ${\boldsymbol \theta}=(\beta_1, \ldots, \beta_{p_1}, c)=:({\boldsymbol \theta}_1, c)$ is assumed $s$-sparse, that is only $s$ entries in ${\boldsymbol \theta}_1$ are different from zero. Here again, like for $W_1$ in ANNs, a non-zero entry in ${\boldsymbol \theta}_1$ corresponds to an entry in the input vector ${\bf x}$ that is relevant to predict the response.
For a properly chosen penalty parameter $\lambda$, LASSO has the remarkable property of retrieving the non-zero entries of ${\boldsymbol \theta}_1$ in certain regimes (that depend on $n$, $p_1$, SNR, training locations~${\cal X}$ and amount $s$ of sparsity); this has been well studied in the noiseless and noisy scenarios by \citet{CandesTao05,DonohoDL06,6034731,BuhlGeer11}, for instance.
In particular, the value of $\lambda$ must bound the sup-norm of the gradient of the empirical loss at zero with high probability when ${\boldsymbol \theta}_1={\bf 0}$ for LASSO to satisfy oracle inequalities.
 For linear models in wavelet denoising theory  \citep{Dono94b}, this approach leads to an asymptotic minimax property. 
Our contribution is to extend the linear methodology to the nonlinear one, and to investigate how well our extension leads to a phase transition to discover underlying nonlinear lower-dimensional structures in the data. 
 

\subsection{Choice of activation functions} \label{subsct:activation}

Since the weights from level two and higher are bounded on the $\ell_2$-ball of unit radius \eqref{eq:Skj}, we require the activation function $\sigma\in {\cal C}^2({\mathbb R})$ to be unbounded. For reasons related to Property~\ref{prop:propconstant} and the choice of the hyperparameter $\lambda$, it must also be null and have a positive derivative at zero:
\begin{equation} \label{sigma(0)=0}
\sigma(0)=0 \quad {\rm and} \quad \sigma'(0)>0.
\end{equation}
The centered {\tt softplus} function $\sigma_{\rm softplus}(u)=\log(1+\exp(u))-\log(2)$  for example satisfies this requirement.
The {\tt ReLU} (Rectified Linear Unit) function $\sigma_{\rm ReLU}(u)=\max(u,0)$ does not because not differentiable at zero.

A legitimate question for a statistician is to ask whether ANNs can retrieve interactions between covariates. Projection pursuit models \citep{FS81} have this ability, which additive models do not have.
For ANNs, owing to their mathematical property of being dense in smooth function spaces, the answer is yes, but with a large number of neurons/parameters when conventional activation functions like {\tt softplus} and {\tt ReLU}  are used.
The following activation functions  (that satisfy the requirements   \eqref{sigma(0)=0}) allow to identify interactions in a sparse way.
\begin{definition}
The smooth activation rescaled dictionary is the collection of activation functions defined by
\begin{equation}\label{eq:sigmafamily}
\sigma_{M,u_0,k}(u)=\frac{1}{k}(f(u)^k-f(0)^k) \quad  {\rm with} \quad f(u)=\frac{1}{M}\log(1+\exp\{M (u+u_0)\})
\end{equation}
indexed by $M>0, u_0>0, k >0$. For $u_0=1$ the dictionary is rescaled in the sense that $\lim_{M\rightarrow \infty}\sigma'_{M,u_0,k}(0)=1$.
\end{definition}
For finite $M$, $\sigma_{M,u_0,j}\in {\cal C}^\infty$. 
For $(u_0,k)=(0,1)$, the family includes two important activation functions: {\tt softplus} for $M=1$ and  {\tt ReLU} as $M$ tends to infinity; with $M=20$, an excellent smooth approximation of ReLU is achieved.

Supposing the association is a single second-order interaction, that is $\mu(x)=x_i x_j$ for some pair $(i,j)$, then $\mu(x)=\mu_\theta^{\rm ANN}(x)$ with
\begin{eqnarray*}
\mu_\theta^{\rm ANN}(x)&=&-1+  \sigma_{\infty,1,2}(x_i+x_j-1)+ \sigma_{\infty,1,2}(-x_i-x_j-1)\\
&&- \sigma_{\infty,1,2}(x_i-1)- \sigma_{\infty,1,2}(-x_i-1)- \sigma_{\infty,1,2}(x_j-1)- \sigma_{\infty,1,2}(-x_j-1)
\end{eqnarray*}
since $x^2/2=1+ \sigma_{\infty,1,2}(x-1)+ \sigma_{\infty,1,2}(-x-1)$. 
When the ANN model employs both linear and quadratic ReLU, selecting neurons  with $\sigma_{\infty,1,k}$ with $k=2$  may reveal interactions between its selected features.
Moreover since $\lim_{M\rightarrow \infty}\sigma'_{M,u_0,k}(0)=u_0^{k-1}$, choosing $u_0=1$ scales all the activation functions in the sense that  their derivatives at zero are asymptotically (as $M\rightarrow \infty$) equal for all~$k$. 

Rescaling allows to mix activation functions with different $k$ in the same ANN; in particular for our choice of hyperparameter $\lambda$, it allows to factorize by $\sigma'(0)$ in Theorem~\ref{th:lambda0} below. Moreover, since sparsity is of interest, zero is a region where the cost function ought to be smooth for optimization purposes; hence choosing $u_0=1$ also makes the wiggliness of the loss function bounded at zero since $\sigma_{M,1,1}''(0)=M/\exp(M)$ while $\sigma_{M,0,1}''(0)=M/4$ (which reflects that ReLU is not differentiable at zero).

\subsection{Selection of penalty $\lambda$}
 \label{subsct:lambda}

The proposed choice of $ \lambda$ is based on Property~\ref{prop:propconstant}.
It shows that fitting a constant function is achieved by choosing $\lambda$ large enough  to set the penalized parameters $ {\boldsymbol \theta}_1$ to zero
 when solving the penalized cost function~\eqref{eq:L1}. For convex loss functions and linear models,
the quantile universal threshold \citep{Giacoetal17} achieves this goal with high probability under the null model that the underlying function is indeed constant.  This specific value  $\lambda_{\rm QUT}$ has good properties for model selection outside the null model as well \citep{Dono94b, Dono95asym}.
The quantile universal threshold has so far been developed and employed for cost functions that are convex in the parameters, hence guaranteeing that any local minimum is also global. The cost function in~\eqref{eq:L1}  is not convex for ANN models, so we extend the quantile universal threshold by guaranteeing with high probability  a local minimum at the sparse point of interest ${\boldsymbol \theta}_{1}={\bf 0}$.
This can be achieved thanks to the penalty term $\lambda\ P({\boldsymbol \theta}_{1})$ that is part of the cost function in~\eqref{eq:L1}, provided $\lambda$ is large enough to create a local minimum with $\hat {\boldsymbol \theta}_1={\bf 0}$.
The following theorem derives an expression for the zero-thresholding function $\lambda_0({\cal Y}, {\cal X})$ which gives the smallest $\lambda$ that guarantees a minimum with   $\hat {\boldsymbol \theta}_1={\bf 0}$, for given output--input data $({\cal Y}, {\cal X})$.
 
 \begin{theorem}
 \label{thm:lambda_choice}
  Consider the optimization problem~\eqref{eq:L1}  with $P({\boldsymbol \theta}_{1})$ defined in~\eqref{eq:Pq} with $q=1$, activation function $\sigma \in {\cal C}^2({\mathbb R})$ and loss function ${\cal L}_n\in {\cal C}^2(\Gamma^n)$  such that $\hat {\bf c}= \arg \min_{ {\bf c}\in {\mathbb R}^m} {\cal L}_n({\cal Y},{\bf c}^{\otimes n})$ exists. Let $ {\boldsymbol \theta}^0=({\bf 0}, W_2 ,  \ldots, W_l,\hat {\bf c})$ with  arbitrary values $W_{k}$  for layers $2$ to $l$.
Define ${ g}_0({\cal Y}, {\cal X}, {\boldsymbol \theta}^0)=\nabla_{{\boldsymbol \theta}_{1}} {\cal L}_n({\cal Y},{\boldsymbol \mu}_{{\boldsymbol \theta}^0}({\cal X}))$.
  If $\lambda > \lambda_0({\cal Y}, {\cal X})=\sup_{(W_2 \ldots, W_{l})}\| { g}_0({\cal Y}, {\cal X}, {\boldsymbol \theta}^0) \|_\infty$, then there is a local minimum  to~\eqref{eq:L1} with $(\hat {\boldsymbol \theta}_{1,\lambda}, \hat {\bf c}_\lambda)=({\bf 0}, \hat {\bf c})$.
 \end{theorem}

 The proof of Theorem~\ref{thm:lambda_choice} is provided in the appendix; it could be made more general for $q\geq 1$ using H{\"o}lder's inequality. 
 In regression for instance, if the loss function between ${\cal Y}\in {\mathbb R}^n$ and $c {\bf 1}$ with $c\in  {\mathbb R}$ and ${\bf 1} \in  {\mathbb R}^n$ is ${\cal L}_n({\cal Y},c{\bf 1}) = \|{\cal Y}-c {\bf 1}\|_2$, then $\hat c=\bar {\cal Y}$,  the average of the responses. Based on~$\lambda_0({\cal Y}, {\cal X})$, the following theorem extends the universal threshold to non-convex cost functions.

 \begin{theorem} \label{th:QUT}
 Given training inputs ${\cal X}$,
 define the random set of outputs 
 ${\cal Y}_0$ generated from from~\eqref{eq:condexpect} with $\mu(\cal X)=\mu_{\boldsymbol \theta}({\cal X})$ defined in~\eqref{eq:muANN} for any activation function satisfying~\eqref{sigma(0)=0}
 under the null hypothesis $H_0: {\boldsymbol \theta}_{1}={\bf 0}$, that is $H_0: \mu_{\boldsymbol \theta}={\bf c}$ is a constant function.
  Letting the random variable $\Lambda=\lambda_0({\cal Y}_0, {\cal X})$ and $F_\Lambda$ be the distribution function of $\Lambda$, the quantile universal threshold is $\lambda_{\rm QUT}=F^{-1}_\Lambda(1-\alpha)$ for a small value of $\alpha$. It satisfies that
\begin{equation}
{\mathbb P}_{H_0}(\mbox{there exists a local minimum to~\eqref{eq:L1} such that } \mu_{\hat {\boldsymbol \theta}_{\lambda_{\rm QUT}}}\mbox { is constant})\geq 1-\alpha.
\end{equation}
 \end{theorem}
 
The law of $\Lambda$ is unknown but can be easily estimated by Monte Carlo simulation, provided there exists a closed form expression for the zero-thresholding function $\lambda_0({\cal Y}, {\cal X})=\sup_{(W_2 \ldots, W_{l})}\| { g}_0({\cal Y}, {\cal X}, {\boldsymbol \theta}^0) \|_\infty$. The following theorem states a simple expression for $\lambda_0({\cal Y}, {\cal X})$ in two important cases: classification and regression.

\begin{theorem} \label{th:lambda0}
 Consider a  fully connected $l$-layer ANN employing a differentiable activation function $\sigma$ and let   $\pi_l=\sqrt{\Pi_{j=3}^l p_{j}}$ for $l\geq 3$, $\pi_2=1$, ${\cal Y}_\bullet={\cal Y}-{\bf 1}_{n}\bar {\cal Y}$ and
 $\|A\|_\infty=\max_{j=1,\ldots,p} \sum_{i=1}^k |a_{ji}|$ for a $p\times k$ matrix $A$.
 \begin{itemize}
  \item In classification, using the cross-entropy ${\cal L}_n ({\cal Y} , {\mu}_{\boldsymbol \theta}( {\cal X}))=\sum_{i=1}^n {\bf y}_i^{\rm T}\log\mu_\theta({\bf x}_i)$ and for the Softmax link function $G$ in~\eqref{eq:GSoft}, we have
  \begin{equation} \label{eq:lambda0C}
\lambda_0({\cal Y}, {\cal X}) =
\pi_l \sigma'(0)^{l-1} \|{\cal X}^{\rm T} {\cal Y}_\bullet \|_\infty;
\end{equation}
\item In regression, for ${\cal L}_n=\| {\cal Y} -  \mu_{ \boldsymbol\theta}({\cal X}) \|_2$,
  we have
  \begin{equation} \label{eq:lambda0G}
\lambda_0({\cal Y}, {\cal X}) =  \pi_l \sigma'(0)^{l-1}
\frac{\|{\cal X}^{\rm T} {\cal Y}_\bullet \|_\infty}{\|{\cal Y}_\bullet\|_2} .
  \end{equation}
 \end{itemize}
\end{theorem}

Theorem~\ref{th:QUT} states that the choice of $\lambda$ is simply an upper quantile of the random variable $\Lambda=\lambda_0({\cal Y}_0, {\cal X})$, where ${\cal Y}_0$ is the distribution of the response under the null distribution that ${\boldsymbol \theta}_1={\bf 0}$. The upper quantile of $\Lambda$
 can be easily estimated by Monte-Carlo simulation.

In regression and assuming Gaussian errors, the null distribution is ${\cal Y}_0 \sim {\rm N}(c {\bf 1}, \xi^2 I_n)$. Both the constant $c$ and $\xi^2$ are unknown however, and $\xi^2$ is difficult to estimate in high dimension. Fortunately, one observes first that \eqref{eq:lambda0G} involves only the mean centered responses ${\cal Y}_\bullet$ and therefore do not dependent on $c$. Second, both numerator and denominator are proportional to $\xi$. Consequently, $\Lambda$ is a pivotal random variable in the Gaussian case. Knowledge of $c$ and $\xi^2$ are therefore not required to derive our choice of hyperparameter $\lambda_{\rm QUT}$. This well-known fact inspired by square-root LASSO \citep{BCW11} motivates the use of ${\cal L}_n=\| {\cal Y} -  \mu_{ \boldsymbol\theta}({\cal X}) \|_2$ rather than ${\cal L}_n=\| {\cal Y} -  \mu_{ \boldsymbol\theta}({\cal X}) \|_2^2$.

In classification, the null distribution is ${\cal Y}_0 \sim {\rm Multinomial}(n, {\bf p}=G({\bf c}))$.  The constant vector ${\bf c}$ is unknown and the random variable $\Lambda$ with $\lambda_0$ defined in~\eqref{eq:lambda0C} is not pivotal. Moreover \citet{Holland73} proved no covariance stabilizing transformation exists for the trinomial distribution. So the approach we take is to assume the training outputs ${\cal Y}$  reflect the proportion of classes in future samples seeking class prediction. So if $\hat {\bf p}$ are the proportions of classes in the training set, then the null distribution is ${\cal Y}_0 \sim {\rm Multinomial}(n, {\bf p}=\hat {\bf p})$. The quantile universal threshold derived under this null hypothesis is appropriate if future data come from the same distribution, which is a reasonable assumption.

 \subsection{Optimization for LASSO ANN}
 \label{subsct:opti}

 For a given $\lambda$, we solve~\eqref{eq:L1}   first by steepest descent with a small learning rate, and then employ a proximal method to refine the minimum by exactly setting to zero some entries  of $\hat W_{1,\lambda_{\rm QUT}}$ \citep{FISTA09,10.1561/2200000015}. 

Solving \eqref{eq:L1}  directly for the prescribed $\lambda=\lambda_{\rm QUT}$ risks getting trapped at some poor local minimum however.
Instead, inspired by simulated annealing and the warm start, we avoid thresholding too hardly at first and possibly missing important features by solving  \eqref{eq:L1} for an increasing sequence of  $\lambda$'s tending to $\lambda=\lambda_{\rm QUT}$, namely $\lambda_k=\exp(k)/(1+\exp(k))\lambda_{\rm QUT}$ for $k\in \{-1,0,\ldots,4\}$. Taking as initial parameter values the solution corresponding to the previous~$\lambda_k$ leads to a sequence of sparser approximating solutions until solving for $\lambda_{\rm QUT}$  at the last step.

The computational cost is low. It requires solving~\eqref{eq:L1} approximately on the small grid of $\lambda$'s tending to $\lambda_{\rm QUT}$ using the warm start to finally solve~\eqref{eq:L1} precisely for $\lambda_{\rm QUT}$.  Calculating $\lambda_{\rm QUT}$ is also cost efficient (and highly parallelizable)  since it is based on an $M$-sample Monte Carlo that calculates $M$ gradients $\{{ g}_0({ y}_k, { X}, {\boldsymbol \theta}_0)\}_{k=1}^M$ using backpropagation \citep{Rumelhart:1986we} for $M$ Gaussian samples $\{y_k\}_{k=1}^M$ under $H_0$.
Using $V$-fold cross-validation instead would require solving ~\eqref{eq:L1} a total of $V*L$ times, where $L$ is the number of $\lambda$'s visited until finding a (hopefully global) minimum to the cross-validation function.
Using a validation set reduces  complexity by a factor $V$, at the cost of using data to validate. 
Instead, our quantile universal threshold approach does not require a validation set.

\section{Regression simulation study}\label{sct:simulation} \label{sct:MCsimu}

The regression problem is model~\eqref{eq:condexpect} for scalar output ($m=1$), Gaussian additive noise and (unknown) standard deviation, here chosen $\xi=1$. 
  To evaluate the ability to retrieve needles in a haystack, 
 the true associations $\mu$ is written as sparse ANNs that uses only $s$ of the $p_1$ entries in the inputs ${\bf x}$.
 We say an association $\mu$ is $s$-sparse when it uses only $s$ input entries, that is  $s=|S|$ where $S=\{j\mid  x_j \mbox{ carries information} \}$ in the association $\mu$.
  A sparse ANN learner estimates which inputs  are relevant by estimating the support with 
  \begin{equation} \label{eq:Shat}
  \hat S=\{j \mid \|\hat {\bf w}_{1,j}\| > \epsilon  \},
  \end{equation}
   where $\hat {\bf w}_{1,j}$ is the $j^\text{th}$ column of the estimated  weights $\hat W_1$ at the first layer. Likewise for linear model~\eqref{eq:lm},  the support is estimated with $\hat S=\{j \mid \hat \beta_j \neq 0\}$.
  
 Since we employ a precise thresholding algorithm to solve~\ref{eq:L1}, we use $\epsilon=0$ to determine $\hat S$ in \eqref{eq:Shat}; other methods aiming at model selection apply a hard thresholding step with a choice for a second hyperparameter $\epsilon$ to get rid of  many small values.
 Our method could be improved by using $\epsilon$ as another hyperparameter,
but our aim is to investigate a phase transition with LASSO ANN, so we consider a single hyperparameter $\lambda$, and show that choosing $\lambda=\lambda_{\rm QUT}$ leads to a phase transition.

 To quantify the performance of the tested methods, we use four criteria: the probability of exact support recovery ${\rm PESR}={\mathbb P}(\hat S=S)$, the true positive rate ${\rm TPR}={\mathbb E}\left (  \frac{|S \bigcap \hat S|}{|S|} \right )$, the false discovery rate ${\rm FDR}={\mathbb E}\left (  \frac{| \bar S \bigcap \hat S|}{|\hat S| \lor 1} \right )$, and the generalization or predictive error ${\rm PE}^2={\mathbb E}(\mu(X)-\hat \mu(X))^2$.
 Although stringent, the PESR criterion reaches values near one in certain regimes.
 In fact, a phase transition has been observed for linear models: PESR is near one  when the complexity parameter $s$ is small, and PESR suddenly decreases to zero when $s$ becomes larger.
One wonders whether this phenomenon is also present for nonlinear models, which we are investigating below. Also, a high TPR with a good control of low FDR is also of interest, but  less strict criteria than high PESR.
Generalization  remains of great concern: ideally a learner should have high TPR and low FDR along with a good generalization performance. 
 
 We consider four learners: a standard ANN with {\tt keras} available in TensorFlow (with its {\tt optimizer=`sgd'} option) with no sparsity inducing mechanism;
  {\tt spinn} (sparse input neural networks) \citep{feng2019sparseinput} with sparsity mechanisms governed by two hyperparameters chosen on a validation set; {\tt spinn-dropout} (which {\tt Python} code was kindly provided to us by the first author)  \citep{pmlr-v80-ye18b} with sparsity inducing mechanisms (including dropout) governed by three hyperparameters chosen on a validation set; and our LASSO ANN with a  sparsity inducing penalty  governed by a single hyperparameter chosen by QUT (i.e., no validation set required).

For LASSO ANN we use two to four-layer ANNs with $(p_2,p_3,p_4)=(20, 10, 5)$, activation function $\sigma_{20,1,1}$ defined in~\eqref{eq:sigmafamily} and the $\ell_1$-LASSO penalty.
{\tt spinn} and {\tt spinn-dropout} use ReLU. 
 The ReLU activation function allows to sparsely write a linear association (Section~ \ref{subsct:lin}) and the nonlinear absolute value function (Section~\ref{subsct:nonlin}).
With Monte-Carlo simulations to estimate PESR, TPR, FDR and PE in two different settings, we compare four learners  as a function of the model complexity parameter $s$, for fixed sample size $n$ and signal to noise ratio governed by $(\xi, \theta)$. 
 The first simulation assumes a sparse linear association and compares LASSO ANN to the benchmark square-root LASSO for linear models.
The second simulation assumes a sparse nonlinear association.
These allegedly simple sparse associations allow to reveal interesting phase transitions in the ability of LASSO ANN to retrieve needles in a haystack. 
Comparing to two other sparsity inducing ANN learners, we observe more coherent phase transitions with LASSO ANN than with the more complex (i.e., more than one hyperparameter) {\tt spinn} and {\tt spinn-dropout} learners in terms of PESR, TPR and FDR.


\subsection{Linear associations} \label{subsct:lin}
 
 The linear model~\eqref{eq:lm} is the most commonly used and studied model, so we investigate in this section how LASSO ANN compares to a state-of-the-art method for linear models, here square-root LASSO  \citep{BCW11} (using the {\tt slim} function in the {\tt flare} library in {\tt R}). This allows to investigate the impact of the loss off convexity for ANNs.
 
Assuming the linear association is  $s$-sparse, this section compares the ability to retrieve the $s$ relevant input entries assuming either a  linear model (the benchmark) or a non-linear model using fully connected ANNs. 
The aim of the Monte Carlo simulation is to investigate:
\begin{enumerate}
 \item a phase transition with LASSO ANN and if so, how close it is to the phase transition of square-root LASSO which, assuming a linear model, should be difficult to improve upon.
 We consider two selection rules for $\lambda$ for square-root LASSO:  QUT and using a validation set to minimize the predictive error. 
 \item how the quantile universal threshold $\lambda_{\rm QUT}$ based  on~\eqref{eq:lambda0G} performs for LASSO ANN with two, three and four layers.
 \item a phase transition with {\tt spinn} and {\tt spinn-dropout}. In an attempt to make them comparable to LASSO, we  set their parameter controlling the trade--off between LASSO and group-LASSO to a small value so that their penalty is essentially LASSO's.
Like LASSO,  {\tt spinn} and {\tt spinn-dropout} use a validation set to tune their hyperparameters. Results with their default values are not as good and not reported here.
\end{enumerate}

This experimental setting allows various interesting comparisons: linear versus nonlinear models to retrieve a linear model, and  model selection- (QUT) versus validation set-based choice of the hyperparameter(s).


We estimate the PESR criterion of the three methods with a Monte-Carlo simulation with $100$ repetitions.
Each sample is generated from an $s$-sparse linear model with $s\in \{0,1,2,\ldots,16\}$, the sample size is  $n=100$ from and the dimension of input variables is $p_1=2n$. 
\citet{preciseunder10} studied in the noiseless case the performance of $\ell_1$-regularization as a function of $\delta=n/p_1$ and $\rho=s/n$ (for us, $\delta=1/2$ and $\rho=s/100$)  and found a PESR phase transition. To be close to their setting, we assume  the input variables are i.i.d.~standard Gaussian with a moderate signal-to-noise ratio:  the $s$ non-zero linear coefficients $\beta_j$ in~\eqref{eq:lm} are all equal to $3$ and the standard deviation of the Gaussian noise is $\xi=1$.
ANN models with ReLU fits linear models sparsely. Indeed a two-layer ANN with a single activated neuron with $s$ non-zero entries in the weights $W_1$  matches the linear function in the convex hull of the data, as stated in the following property.

\begin{property} 
 \label{property:linearANN}
Using the ReLU activation function, an $s$-sparse linear function restricted to the convex hull of the $n$ data vectors $\{{\bf x}_i\}_{i=1,\ldots,n}$ can be written as a two-layer neural network with a single neuron with a row matrix $W_1$  with $s$ non-zero entries.
\end{property} 

The proof of Property~\ref{property:linearANN} is provided in the appendix. 
The convex hull includes the $n$ observed covariates which enter the square-root $\ell_2$-loss in~\eqref{eq:L1}. So the sparsest two-layer ANN model that solves the optimization and that is a linear model in the convex hull of the data has a single neuron. But the ANN fit is no longer linear outside the convex hull, which makes prediction error PE poor outside the convex hull range of the data; we therefore do not report PE for the linear model since the ANN model will have poor performance for test data outside the convex hull of the training data.

\begin{figure}
\includegraphics[width=6.4in]{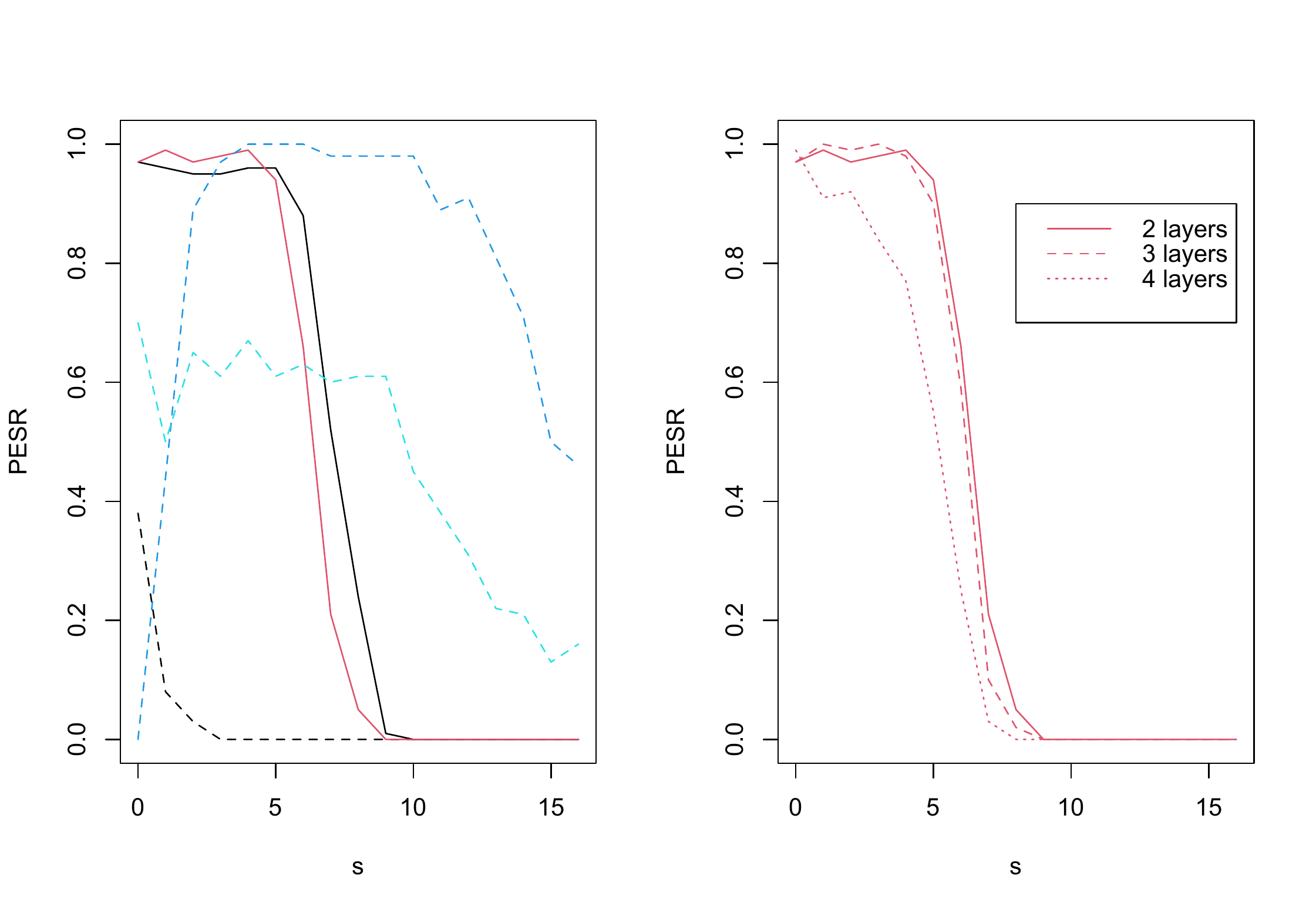}
\caption{Monte-Carlo simulation results for linear association plotting the estimated probability of exact support recovery (PESR).
Left plot: the two black curves assume a linear model while the color curves assume an ANN model; the two blue lines (light for {\tt spinn} and dark for {\tt spinn-dropout}) are governed by more than one hyperparameter while the red line (LASSO ANN) is governed by a single hyperparameter;
the two continuous lines (black for square-root LASSO linear and red for LASSO ANN) select the hyperparameter with QUT while the dashed lines require a validation set.
Right plot: LASSO ANN with 2 to 4 layers with its hyperparameter based on QUT.} 
\label{fig:linear}
\end{figure}

Figure~\ref{fig:linear} summarizes the results of the Monte-Carlo simulation. As in \citet{preciseunder10}, we observe a PESR phase transition. Surprisingly, little is lost with LASSO ANN (red curve) compared to linear model based on QUT (black line), showing the good performances of both the choice of $\lambda_{\rm QUT}$ and the optimization employed for LASSO ANN. 
Linear model based on a validation set (black dashed line) shows poor performance in terms of PESR, as expected.
In summary, LASSO ANN compares surprisingly well  to the benchmark linear square-root LASSO with QUT by not losing much in terms of PESR.
The other two ANNs learners  {\tt spinn} and {\tt spinn-dropout} cannot directly be compared to the other since governed by more than one hyperparameter, but, while we observe good PESR for $s$ large, their global behavior does not follow the conventional phase transition (that is, no high plateau near one for small $s$ and rapidly dropping down to zero with larger~$s$); the nonlinear simulation of next Section also reveals some non-conventional behaviors for these two ANN learners.
Going back to LASSO ANN, we observe on the right plot of Figure~\ref{fig:linear} that using more layers slightly lowers the performance, as expected, but that the choice of $\lambda_{\rm QUT}$ for more layers still leads to a conventional phase transition.

 \subsection{Nonlinear associations} \label{subsct:nonlin}

To investigate a phase transition for nonlinear sparse associations, we consider  $s$-sparse functions of the form
$
\mu_{\boldsymbol{\theta}}({\bf x})=  \sum_{i=1}^h 10\cdot |x_{2i}-x_{2i-1}|  
$
for $h \in \{0,1,\ldots,8 \}$, which corresponds to $s$ needles in a nonlinear haystack with $s\in \{ 0, 2, \ldots, 16 \}$.
Because this association is harder to retrieve than the linear one (due to the non-monotone nature of the absolute value function),  the haystack is of size $p_1=50$ and the training set is of size $n=500$. This ratio $\delta=n/p_1=10$ seems to be the limit where needles can be recovered with LASSO ANN.
The association
$\mu_{\boldsymbol{\theta}}({\bf x})$ is well approximated by a sparse two-layer ANN employing the smooth activation function $\sigma_{20,1,1}$ and
with $c=10s$, ${\bf w}_{2}=(10\cdot {\bf 1}_{h}^{T}$, ${\bf 0}_{p_2/2-h}^{T},10\cdot {\bf 1}_{h}^{T}$, ${\bf 0}_{p_2/2-h}^{T})$, ${\bf b}_{1}=-{\bf 1}_{p_2}$ and
\newcommand\coolrightbrace[2]{%
\left.\vphantom{\begin{matrix} #1 \end{matrix}}\right\}#2}
\begin{equation}
W_1=\left [ \begin{array}{c}
W \\ W 
\end{array}
\right ]
\ {\rm with} \ 
W =\left [ \begin{array}{cccccccccccccc}
-1 & 1 & 0 &  0 &  \ldots &  & & &  & &\ldots & 0\\
 0 & 0 & -1 &  1 &  0 &  \ldots & & & & &  \ldots& 0\\
  \vdots & &&&& \vdots &\vdots&&& && &  \vdots \\
0 & \ldots &  & & \ldots&  0 & -1 & 1 & 0 &  \ldots & \ldots &  0 \\
0 & \ldots & &&& \ldots &&& & &\ldots & 0 \\
\vdots &  &&&&&&&&&& \vdots \\ 
0 & \ldots &  &&& \ldots &&&&&\ldots& 0\\
\end{array}
\right ]
\begin{matrix}
    \coolrightbrace{0 \\ 0 \\ \vdots \\ 0}{h\hspace{21.5pt}}\\
    \coolrightbrace{0 \\ \vdots \\ 0 }{\frac{p_2}{2}-h}\\
\end{matrix}.
\end{equation}
The columns of $W_1$ being sparse, a LASSO penalty is more appropriate than a group-LASSO penalty.

\begin{figure}
\includegraphics[width=6.4in]{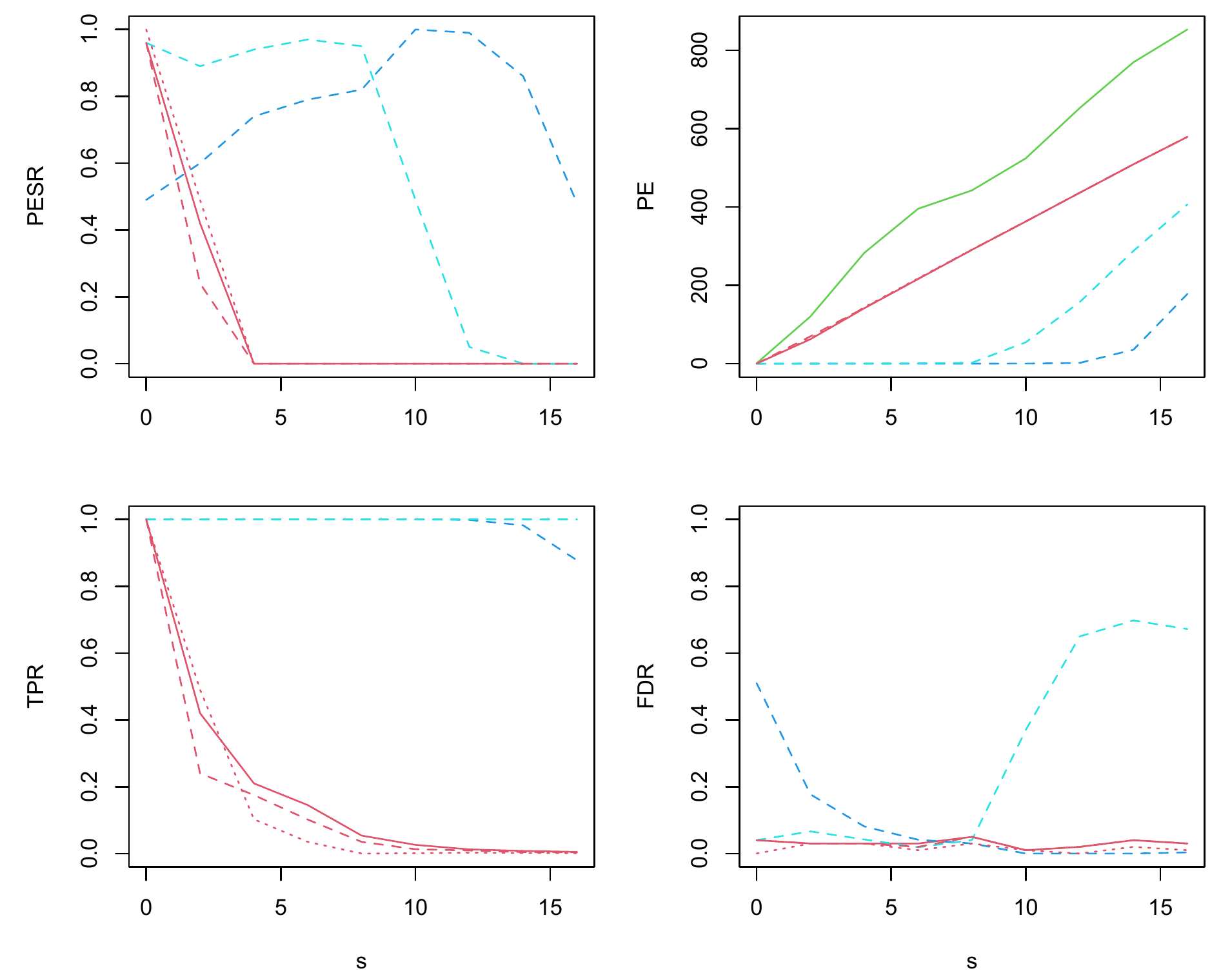}
\caption{Monte-Carlo simulation results for nonlinear association plotting the estimated probability of exact support recovery (PESR -- top left), generalization performance (PE -- top right),
true positive rate (TPR -- bottom left) and false discovery rate (FDR -- bottom right).
The red curves are for LASSO ANN with its hyperparameter based on QUT with two (continuous) to four layers (dashed).
The two blue lines (light for {\tt spinn} and dark for {\tt spinn-dropout}) are governed by more than one hyperparameter selected based on a validation set. The green curve is a standard ANN (without sparsity constraint).} 
\label{fig:nonlinear}
\end{figure}

  Figure~\ref{fig:nonlinear} reports the estimated  PESR, TPR, FDR and PE criteria as a function of the sparsity level $s$. 
We observe that, as for linear models,  LASSO ANN (red lines for two to four layers) has a PESR phase transition thanks to a good trade--off between high TPR and low FDR. Moreover LASSO ANN has better generalization performance in this setting than the off-the-shelf ANN learner (green lines). 
The other two ANNs learners  {\tt spinn} and {\tt spinn-dropout} (light and dark blue, respectively) perform somewhat better in terms of PESR thanks to more than one hyperparameter, but not with a monotone way for {\tt spinn-dropout}; moreover, the FDR of  {\tt spinn} and {\tt spinn-dropout}  is not well controlled along the sparsity range indexed by $s$. The good FDR control of LASSO ANN is striking, in particular at $s=0$ where its value is near $\alpha=0.05$, as expected, proving the effectiveness of not only QUT but also of the optimization algorithm.
Finally, as far as generalization is concerned, the sparsity inducing learners perform better than the conventional ANN learner since the underlying ANN model is indeed sparse. Because LASSO ANN not only selects a sparse model but also shrinks, its predictive performance is not as good as with {\tt spinn} and {\tt spinn-dropout} which  regularization parameters are selected to generalize well.

\subsection{Conclusions of the Monte Carlo simulations}

With a single hyperparameter, LASSO ANN  has a phase transition for both linear and nonlinear associations and a good FDR control.
This reveals that the quantile universal threshold and the optimization scheme employed are performant.
With the linear simulation, we observe that the impact of the loss off convexity is mild with LASSO ANN since we essentially get the same phase transition as with a linear model.
The other ANN learners considered do not have a conventional phase transition and do not control their FDR well; yet, with the help of more hyperparameters, they are able to generalize well.




 \section{Application to real data} \label{sct:appli}
 
 \subsection{Classification data} \label{subsct:classif}
 
The characteristics of 26 classification data sets are listed in Table~\ref{tab:classification-data}, in particular the sample size $n$, the number of inputs $p_1$ and the number of classes $m$. Most inputs are gene expressions, but there are also FFT preprocessed time series  and other types of inputs.
 
\begin{figure}
\includegraphics[width=6.5in, height=7.5in]{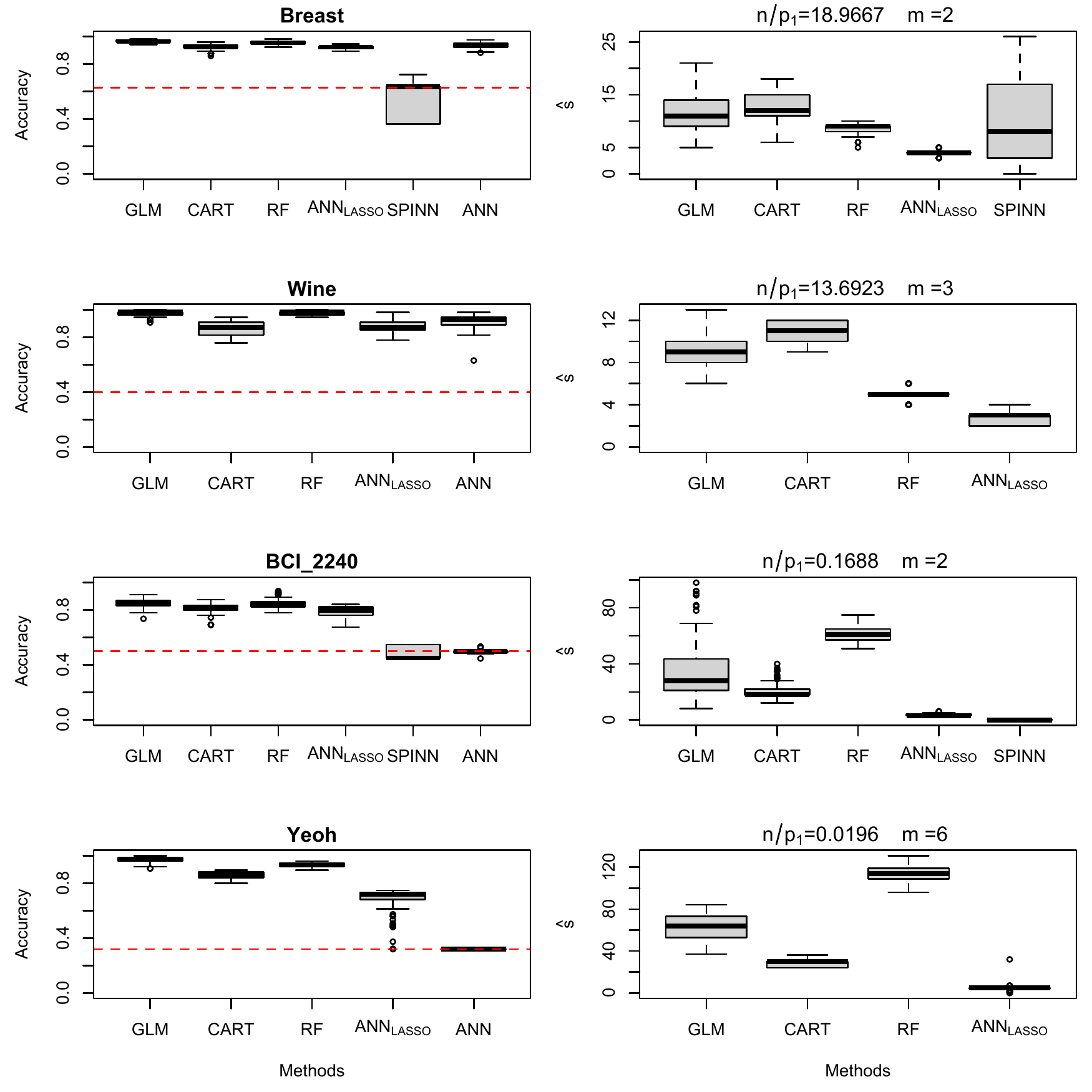}
\caption{Monte-Carlo simulation results based on four representative data sets, namely, Breast, Wine, BCI$\_$2240, Yeoh  of Table~\ref{tab:classification-data}. The left boxplots are the accuracy results and the left boxplots are the number of selected needles. The horizontal red line is the accuracy by always predicting the most frequent class (that is, without looking at the inputs).} 
\label{fig:classif}
\end{figure}

\begin{center}
\begin{figure}[h]
\includegraphics[width=6.5in, height=4.0in]{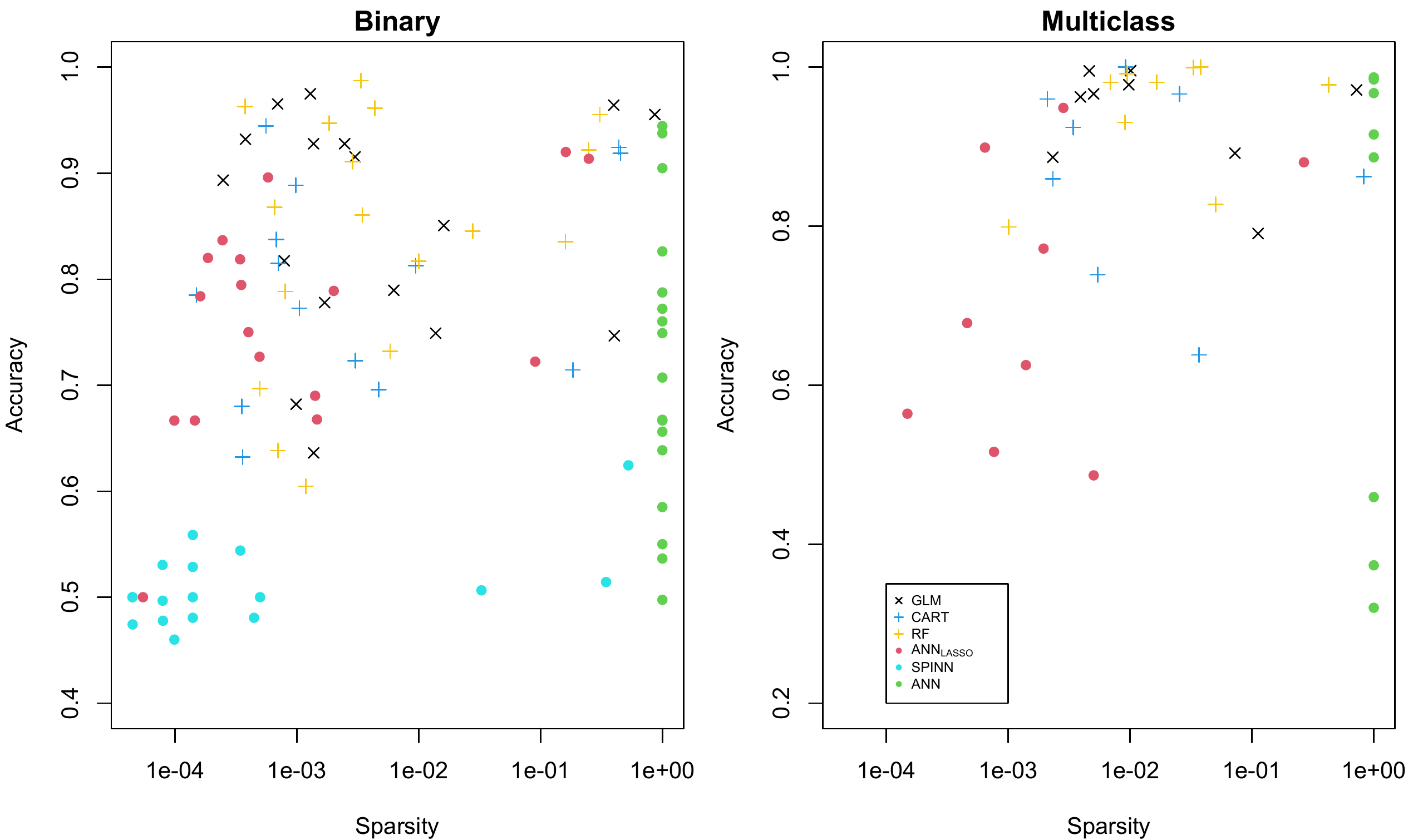}
\caption{Summary of Monte-Carlo results for all data sets of Table~\ref{tab:classification-data}. The $x$-axis measures sparsity on a log-scale with $\log((\hat s+1)/(p_1+1))$ and the $y$-axis is accuracy.} 
\label{fig:scattplot}
\end{figure}
\end{center}

 We randomly split the data into training (70\%) and test (30\%) sets, repeating the operation 100 times.  Figure~\ref{fig:classif} reports the results  for four data sets chosen for their ratios $n/p_1$ and their number of classes $m$ (marked with a $\dagger$ in Table~\ref{tab:classification-data}). The left boxplots of Figure~\ref{fig:classif} report classification accuracy, and the right boxplots report the number of selected needles $\hat s$.  High accuracy along with low $\hat s$ reflects good needles selection.
 The results of the remaining 22 sets are plotted in the scatter plot of Figure~\ref{fig:scattplot}.

 We train and test the following learners: LASSO GLM  with $\lambda$ chosen to minimize $10$-fold cross validation \citep{GLMnet} in {\tt R} with {\tt glmnet},
CART \citep{cart84} in {\tt R} with {\tt rpart}, random forest \citep{breiman2001random} in {\tt R} with {\tt randomForest}, SPINN in {\tt Python} for binary classification ({\tt https://github.com/jjfeng/spinn}; no code for multiclass and for {\tt spinn-dropout} available), standard ANN learner in {\tt Python}  with {\tt keras} and its {\tt optimizer=`adam'} option, and our LASSO ANN with two layers in {\tt Python}. For random forest, there is no clear way of counting the number of needles, but we choose to select as needles those inputs which corresponding p-values (provided by {\tt randomForestExplainer}) are smaller than $\alpha=0.05$ after a Bonferoni adjustment. Random forest is an ensemble learner that combines CARTs; so the comparison between CART and random forest quantifies the improvement achieved by ensembling learners, and the comparison between CART and LASSO ANN is more fair since both are no ensemble learners.

\tiny
\begin{table}[h!]
\caption{Some data characteristics (results for data with $\dagger$ are plotted in Figure~\ref{fig:classif}).}
\begin{center}
\resizebox{1\columnwidth}{!}{
\begin{tabular}{llrrrrl}
\toprule
{\bf Dataset} & {\bf Domain} & {\bf n} & ${\bf p}_1$ &{\bf n}/${\bf p}_1$& {\bf m}&{\bf Source} \\
\midrule 
Climate &Climate model&540&18&30.0000&2&UCI-MLR\\
Breast$\dagger$ &Breast cancer &569&30&18.9667&2&python sklearn\\
Wine$\dagger$&Wine&178&13&13.6923&3&python sklearn\\
Connectionist&Connectionism&208&60&3.4667&2&UCI-MLR\\
Bearing&Engine noise&952&1024&0.9297&4&CWRU data center\\
Sorlie&Breast cancer&85&456&0.1864&5&R: datamicroarray\\
BCI$\_2240^\dagger$&Brain signal&378&2240&0.1688&2& BCI competition \\
Christensen&Medical&217&1413&0.1536&3&R: datamicroarray\\
Genes&Cancer RNA&801&12356&0.0648&5&UCI-MLR\\
Gravier&Breast cancer&168&2905&0.0578&2&R: datamicroarray\\
Alon&Colon cancer&62&2000&0.0310&2&R: datamicroarray\\
Khan&Blue cell tumors&63&2308&0.0273&4&R: datamicroarray\\
Yeoh$\dagger$&Leukemia&248&12625&0.0196&6&R: datamicroarray\\
Su&Medical&102&5565&0.0183&4&R: datamicroarray\\
Gordon&Lung cancer&181&12533&0.0144&2&R: datamicroarray\\
Tian&Myeloma&173&12625&0.0137&2&R: datamicroarray\\
Shipp&Lymphoma&77&7129&0.0108&2&R: datamicroarray\\
Golub&Leukemia&72&7129&0.0101&2&R: datamicroarray\\
Pomeroy&Nervous system&60&7128&0.0084&2&R: datamicroarray\\
Singh&Prostate cancer&102&12600&0.0081&2&R: datamicroarray\\
West&Breast cancer&49&7129&0.0069&2&R: datamicroarray\\
Burczynski&Crohn's disease&127&22283&0.0057&3&R: datamicroarray\\
Chin&Breast cancer&118&22215&0.0053&2&R: datamicroarray\\
Subramanian&Medical&50&10100&0.0050&2&R: datamicroarray\\
Chowdary&Breast cancer&104&22283&0.0047&2&R: datamicroarray\\
Borovecki&Medical&31&22283&0.0014&2&R: datamicroarray\\
\bottomrule
\end{tabular}
}
\end{center}
\label{tab:classification-data} 
\end{table}

\normalsize

Figure~\ref{fig:scattplot} visualizes the sparsity--accuracy trade--off  by plotting accuracy versus $\log(a \hat s/p_1+1)$ with $a=\exp(1)-1$, so that both axes are on $[0,1]$. Learners with points near $(0,1) $ offer the best trade--off. Left is for binary and right  for multiclass classifications.
Among all ANN-based learners (represented with ``o''), LASSO ANN is clearly the best.

The main lesson of this experiment on real data sets is that LASSO ANN offers a good compromise between high accuracy and low number of selected needles. Yet, linear learners are difficult to beat when $n/p_1\ll 1$, which corroborates our findings in regression that the sample size must be large to identify nonlinear associations.

\subsection{Regression data} \label{subsct:regression}
 
\citet{PeterBulbiology:14} reported genetic data measuring  the  expression  levels  of $p_1=  4088$  genes on $n=  71$  Bacillus  subtilis  bacteria.   The logarithms of gene expression measurements are known to have some strongly correlated genes, which also makes selection difficult.  The output is the riboflavin production rate of the bacteria. This is a high-dimensional setting in the sense that the training set is small  compared to the size of the haystack. Generalization is not the goal here, but finding the informative genes; the scientific questions are: what genes affect the riboflavin production rate? Is the association linear?
 
The ground truth is not known here, but LASSO-zero, a conservative method with low false discovery rate \citep{DesclouxSardy2018}, selects genes   $4003$ and $2564$. Standard LASSO (using {\tt cv.glmnet}  in {\tt R}) selects 30 genes including $4003$ and $2564$.
Using $p_2=20$ neurons, LASSO ANN  finds a single active neuron containing 9 non-zero parameters including genes $4003$ and $2564$.
\citet{feng2019sparseinput} reports 45 important genes with {\tt spinn}, and running {\tt spinn-dropout} 100 times (randomly splitting into $70\%$ training and $30\%$ validating) we find an average of 6 genes (in which $4003$ and $2564$ are rarely present).  
 So the answers to the scientific questions are that few genes seem  responsible for riboflavin production  and that a linear model seems sufficient (a single neuron is active).
 
 
 \section{Conclusion} \label{sct:conclusion}

For finding needles in a nonlinear haystack, LASSO ANN is an artificial neural networks learner that, with a simple principle to select a single hyperparameter, achieves: (1)  a phase transition in the probability of exact support recovery and controls well the false discovery rate; (2) a consistent good trade--off between generalization and low number of selected needles whether in regression, binary or multiclass classification or various $n/p_1$ ratios. This makes it a good candidate to discover important features without too many spurious ones.
Our empirical findings call for more theory to mathematically predict the regimes indexed by  $(n,{\bf p},s,\xi,\theta,\sigma )$ where feature recovery is highly probable. We also introduced a class of rescaled activation functions $\sigma_{M,u_0,k}$ that can be employed within the same ANN model, for instance to fit interactions in a sparse way.

ANN models are widely used state-of-the-art black boxes.  There is a keen interest, especially in scientific  applications, to understand the why of model predictions.
 Sparse encoding automatic feature selection provides a path towards such an understanding.
   Our work makes sparse encoding with LASSO ANN closer to practical applications. Its coherent PESR behavior and FDR control make it reliable for finding needles in nonlinear haystacks, but could also be used for other ANN tasks requiring sparsity, e.g., sparse auto-encoding or convolutional ANN \citep{cnn2020}. Inspired by {\tt spinn-dropout}, the idea of pruning \citep{8578988,ChaoWXC20} and more generally of subset selection that preceded LASSO, we could still improve LASSO ANN with dropout.


   \section{Reproducible research}
   
   Our codes are available at \href{https://github.com/StatisticsL/ANN-LASSO}{https://github.com/StatisticsL/ANN-LASSO}.
   
 \section{Acknowledgments}

The first author has been supported in Switzerland by China Scholarship Council, Award Number 202006220228.
Yen Ting Lin and Nick Hengardner have been supported by the Joint Design of Advanced Computing Solutions for Cancer program established by the U.S.~Department of Energy and the National Cancer Institute of the National Institutes of Health under Contract DE-AC5206NA25396 and Laboratory Directed Research and Development program under project number 20210043DR (Uncertainty Quantification for Robust Machine Learning). We thank Professor Mao Ye for providing us with the {\tt spinn} and {\tt spinn-dropout} Python codes, and Dr.~Thomas Kerdreux and Mr.~Pablo Strasser for their help with Python.

\appendix

\section{Proof of Theorem 1}
\begin{proof}
	Let ${W}_1$ be any matrix with $\norm{{W}_1}_1 = 1$ and ${\bf c}_1\in \mathbb{R}^r$ be any vector with $\norm{{\bf c}_1}_1 = 1$. Let ${{\boldsymbol \theta}^\epsilon}
	=(\epsilon W_1, W_2, \ldots, W_{l},\hat {\bf c}+\epsilon {\bf c}_1)$ for any $\tilde {\boldsymbol \theta}=(W_2,  \ldots, W_{l})$.
	Since the loss function $l$ is twice differentiable with respect to $W_1$ around ${\boldsymbol \theta}^0$,
	applying Taylor's theorem we have
	\begin{eqnarray*}
		|{\cal L}_n({\cal Y}, \mu_{{\boldsymbol \theta}^\epsilon}({\cal X})) - {\cal L}_n({\cal Y}, \mu_{{\boldsymbol \theta}^0}({\cal X}))| &=& |\nabla_{W_1} {\cal L}_n ({\cal Y}, \mu_{{\boldsymbol \theta}^0}(x)) (\epsilon {W}_1)+\nabla_{{\bf c}_1} {\cal L}_n({\cal Y}, \mu_{{\boldsymbol \theta}^0}({\cal X})) (\epsilon {\bf c}_1)\\
		&&+ o(\epsilon^2 \norm{{W}_1}_1)+ o(\epsilon^2 \norm{{\bf c}_1}_1)| \\
		&=& |\epsilon g_0({\cal Y},{\cal X}, {\boldsymbol \theta}^0) {W}_1 +0+ o(\epsilon^2) |\\
		&\leq& |\epsilon| \norm{g_0({\cal Y},{\cal X},{\boldsymbol \theta}^0)}_\infty + o(\epsilon^2).
	\end{eqnarray*}
	Therefore we get
	$
		{\cal L}_n({\cal Y}, \mu_{{\boldsymbol \theta}^\epsilon}({\cal X})) \geq {\cal L}_n({\cal Y}, \mu_{{\boldsymbol \theta}^0}({\cal X})) - |\epsilon| \norm{g_0({\cal Y},{\cal X},{\boldsymbol \theta}^0)}_\infty + o(\epsilon^2).
	$
	Since we are looking for a universal $\lambda$ we  take the supremum over all valid choices of these layers. Since we normalize these layers, the supremum is finite. If we now assume that $(\lambda - \sup_{\tilde {\boldsymbol \theta}}\norm{g_0({\cal Y},{\cal X},{\boldsymbol \theta}^0)}_\infty) > C > 0$ it follows for the regularized loss function that
	\begin{align*}
		{\cal L}_n({\cal Y}, \mu_{{\boldsymbol \theta}^\epsilon}({\cal X})) + \lambda \norm{\epsilon {W}_1}_1 &\geq {\cal L}_n({\cal Y}, \mu_{{\boldsymbol \theta}^0}({\cal X})) + |\epsilon| (\lambda - \norm{g_0({\cal Y},{\cal X},{\boldsymbol \theta}^0)}_\infty) + o(\epsilon^2) \\
		&> {\cal L}_n({\cal Y}, \mu_{{\boldsymbol \theta}^0}({\cal X})) + |\epsilon| C + o(\epsilon^2) \\
		&> {\cal L}_n({\cal Y}, \mu_{{\boldsymbol \theta}^0}({\cal X})) \quad \text{for $|\epsilon|$ small enough.}
	\end{align*}
	Thus the cost function~\eqref{eq:L1} with our choice of $\lambda$ indeed has a local minimum at ${\boldsymbol \theta}^0$.
\end{proof}

\section{Proof of Theorem ~\ref{th:lambda0}}

In regression, the square-root $\ell_2$-loss is
\begin{equation*}
\begin{split}
{\cal L}_n({\cal Y}, \mu_{\boldsymbol{\theta}}({\cal X}))&=\| {\cal Y} -  \mu_{ \boldsymbol\theta}({\cal X}) \|_2\\
&=\sqrt{
	\sum_{k=1}^{n}\left(y_k-(c+{\bf w}_{l}\sigma({\bf b}_{l-1}+W_{l-1}\sigma(\cdots\sigma({\bf b}_{1}+W_1\boldsymbol{x}_k)))
	\right)^{2}},
\end{split}
\end{equation*}
where $y_k \in \mathbb{R}$, $c \in \mathbb{R}$, ${\bf w}_l \in 1 \times \mathbb{R}^{p_l}$, $W_{k} \in \mathbb{R}^{p_{k+1} \times p_{k}},k=1,\cdots,l-1$ and ${\bf b}_{k}\in \mathbb{R}^{p_{k+1} \times 1},k=1,\cdots,l-1$.
At ${\boldsymbol{\theta}_1}={\bf 0}$, the least squares problem is solved for $\hat{c} = \bar{{\cal Y}}$, the average of the training set responses. So we want to evaluate the gradient with respect to ${\boldsymbol{\theta}_1}$ at $({\boldsymbol{\theta}_1}, c)=({\bf 0}, \bar {\cal Y})$, which we call condition $(L_0)$. Let us consider the partial derivative of every entry. Some elementary calculation yields
\begin{equation*}
\begin{split}
\frac{\partial {\cal L}_n({\cal Y}, \mu_{\boldsymbol{\theta}}({\cal X}))}{\partial W_{1;i,j}}\bigg|_{L_0}&=
\frac{\sigma^{'}(0)^{l-1}}{\|{\cal Y}_\bullet\|_2} \left(-{\bf w}_lW_{l-1}\cdots W_{2;:i}\right)\sum_{k=1}^{n}(y_k-\bar{{\cal Y}})x_{k,j},
\end{split}
\end{equation*}	
where $W_{2;:i}$ is the i-th column with $W_{2}$. And
\begin{equation*}
\begin{split}
\frac{\partial {\cal L}_n({\cal Y}, \mu_{\boldsymbol{\theta}}({\cal X}))}{\partial b_{h;i}}\bigg|_{L_0}&=
\frac{\sigma^{'}(0)^{l-h}}{\|{\cal Y}_\bullet \|_2} \left(-{\bf w}_3\cdots W_{h+1;:i}\right)\sum_{k=1}^{n}(y_k-\bar{{\cal Y}}),
1\leq h \leq (l-2)
\end{split}
\end{equation*}	
\begin{equation*}
\begin{split}
\frac{\partial {\cal L}_n({\cal Y}, \mu_{\boldsymbol{\theta}}({\cal X}))}{\partial b_{l-1;i}}\bigg|_{L_0}&=
\frac{\sigma^{'}(0)}{\|{\cal Y}_\bullet\|_2} \left(-{\bf w}_{l;i}\right)\sum_{k=1}^{n}(y_k-\bar{{\cal Y}}).
\end{split}
\end{equation*}	
Now,
\begin{equation*}
\nabla_{\boldsymbol{\theta}_1}{{\cal L}_n({\cal Y}, \mu_{\boldsymbol{\theta}^{0}}({\cal X}))}
=\begin{pmatrix}
\frac{\sigma^{'}(0)^{l-1}}{\|{\cal Y}_\bullet\|_2} \left(-{\bf w}_lW_{l-1}\cdots W_{2;:i}\right)\sum_{k=1}^{n}(y_k-\bar{{\cal Y}})x_{k,1}\\
\vdots\\
\frac{\sigma^{'}(0)^{l-1}}{\|{\cal Y}_\bullet\|_2} \left(-{\bf w}_lW_{l-1}\cdots W_{2;:i}\right)\sum_{k=1}^{n}(y_k-\bar{{\cal Y}})x_{k,p_1}\\
\frac{\sigma^{'}(0)^{l-1}}{\|{\cal Y}_\bullet\|_2} \left(-{\bf w}_3\cdots W_{2;:i}\right)\sum_{k=1}^{n}(y_k-\bar{{\cal Y}})
\\
\vdots\\
\frac{\sigma^{'}(0)^{2}}{\|{\cal Y}_\bullet\|_2} \left(-{\bf w}_3W_{l-1;:i}\right)\sum_{k=1}^{n}(y_k-\bar{{\cal Y}})
\\
\frac{\sigma^{'}(0)}{\|{\cal Y}_\bullet\|_2} \left(-{\bf w}_{l;i}\right)\sum_{k=1}^{n}(y_k-\bar{{\cal Y}})
\end{pmatrix}.
\end{equation*}
Since $\sum_{k=1}^{n}(y_k-\bar{{\cal Y}})=0$, many entries of the gradient are zero.
Via choosing 
$$
W_{2}=\begin{pmatrix}{}
  1&0&\cdots&0 \\
  \vdots& \vdots&\ddots&\vdots \\
  1&0&\cdots&0
\end{pmatrix},
\quad
W_{k}=\begin{pmatrix}{}
  \frac{1}{\sqrt{p_k}}&\frac{1}{\sqrt{p_k}}&\cdots&\frac{1}{\sqrt{p_k}} \\
  \vdots& \vdots&\ddots&\vdots \\
  \frac{1}{\sqrt{p_k}}&\frac{1}{\sqrt{p_k}}&\cdots&\frac{1}{\sqrt{p_k}}
\end{pmatrix} 
\text{for }k=3,\cdots,l-1, 
$$
and
${\bf w}_{l}^{\rm T}=\frac{1}{\sqrt{p_l}} {\bf 1}$, we have
\begin{equation*}
\begin{split}
\sup_{\begin{subarray}{c}
\|W_h;i:\|_2=1,\\
h=2,\cdots,l-1,\\
i=1,\cdots,p_{h+1};\\
\|{\bf w}_l\|_2=1.
\end{subarray}	
}
\left\| \nabla_{\boldsymbol{\theta}_1}{{\cal L}_n({\cal Y}, \mu_{\boldsymbol{\theta}^{0}}({\cal X}))}\right\|_{\infty}
&=\max_{i,j}\left\{
\frac{\sigma^{'}(0)^{l-1}}{\|{\cal Y}_\bullet\|_2} \left(-{\bf w}_lW_{l-1}\cdots W_{2;:i}\right)\sum_{k=1}^{n}(y_k-\bar{{\cal Y}})x_{k,j}\right\}\\
&=\frac{\sqrt{p_3\cdots p_l}\sigma^{'}(0)^{l-1}}{\|{\cal Y}_\bullet\|_2}\left\|\sum_{k=1}^{n}(y_k-\bar{{\cal Y}})\boldsymbol{x}_{k}\right\|_{\infty}\\
&=\pi_l \sigma'(0)^{l-1}
\frac{\|{\cal X}^{\rm T} {\cal Y}_\bullet \|_\infty}{\|{\cal Y}_\bullet\|_2}, 
\end{split}
\end{equation*}
the closed form expression for $\lambda_0({\cal Y}, {\cal X})$  of Theorem~\ref{th:lambda0} for regression.

In classification, the cross-entropy loss is
\begin{equation*}
\begin{split}
{\cal L}_n({\cal Y}, \mu_{\boldsymbol{\theta}}({\cal X}))&=-\sum_{k=1}^{n}\boldsymbol{y}_{k}^{T}\log\hat{\boldsymbol{p}}_{k}\\
&=-\sum_{k=1}^{n}\sum_{t=1}^{m}y_{k,t}\left[c_t+W_{l;t:}\sigma({\bf b}_{l-1}+W_{l-1}\sigma(\cdots\sigma({\bf b}_{1}+W_1\boldsymbol{x}_k)))\right.\\
&\left.-\log\left\{\sum_{h=1}^{m}\exp\left\{c_h+W_{l;h:}\sigma({\bf b}_{l-1}+W_{l-1}\sigma(\cdots\sigma({\bf b}_{1}+W_1\boldsymbol{x}_k)))\right\}\right\}\right],
\end{split}
\end{equation*}
where $W_{l;h:}$ is the h-th row in $W_{l}$.
The derivatives with respect to every element are:
\begin{equation*}
\begin{split}
\frac{\partial {\cal L}_n({\cal Y}, \mu_{\boldsymbol{\theta}}({\cal X}))}{\partial W_{1;i,j}}\bigg|_{L_0}&=
\sigma^{'}(0)^{l-1}\sum_{k=1}^{n}x_{k,j}\left(\bar{{\cal Y}}-\boldsymbol{y}_k\right)^{T}W_l \cdots W_{2;:i},
\end{split}
\end{equation*}	
\begin{equation*}
\begin{split}
\frac{\partial {\cal L}_n({\cal Y}, \mu_{\boldsymbol{\theta}}(\boldsymbol{x}))}{\partial b_{h;i}}\bigg|_{L_0}&=
\sigma^{'}(0)^{l-h}\sum_{k=1}^{n}\left(\bar{{\cal Y}}-\boldsymbol{y}_k\right)^{T}W_l\cdots W_{h+1;:i}, \text{ } 1 \leq h \leq l-2,
\end{split}
\end{equation*}	
\begin{equation*}
\begin{split}
\frac{\partial {\cal L}_n({\cal Y}, \mu_{\boldsymbol{\theta}}(\boldsymbol{x}))}{\partial b_{l-1;i}}\bigg|_{L_0}&=
\sigma^{'}(0)\sum_{k=1}^{n}\left( \bar{{\cal Y}}-\boldsymbol{y}_k\right)^{T}W_{l;:i},
\end{split}
\end{equation*}	
leading to
\begin{equation*}
\nabla_{\boldsymbol{\theta}_1} {\cal L}_n({\cal Y}, \mu_{\boldsymbol{\theta}^{0}}(\boldsymbol{x}))=\begin{pmatrix}
\sigma^{'}(0)^{l-1}\sum_{k=1}^{n}x_{k,1}\left(\bar{{\cal Y}}-\boldsymbol{y}_k\right)^{T}W_l \cdots W_{2;:i}\\
\vdots\\
\sigma^{'}(0)^{l-1}\sum_{k=1}^{n}x_{k,p_1}\left(\bar{{\cal Y}}-\boldsymbol{y}_k\right)^{T}W_l \cdots W_{2;:i}\\
\sigma^{'}(0)^{l-1}\sum_{k=1}^{n}\left(\bar{{\cal Y}}-\boldsymbol{y}_k\right)^{T}W_l\cdots W_{2;:i}\\
\vdots\\
\sigma^{'}(0)^{2}\sum_{k=1}^{n}\left(\bar{{\cal Y}}-\boldsymbol{y}_k\right)^{T}W_lW_{l-1;:i}\\
\sigma^{'}(0)\sum_{k=1}^{n}\left( \bar{{\cal Y}}-\boldsymbol{y}_k\right)^{T}W_{l;:i}
\end{pmatrix}.
\end{equation*}
Via choosing 
$W_{2;:i}=[1,1,\cdots,1]_{p_3 \times 1}^{T}$, $W_{h_j;i:}^{T}=1/\sqrt{p_h} {\bf 1}$
 for  $h=3,\cdots,l-1; i=1,2,\cdots,p_{h+1}$, and
$$
W_{l_j;i:}^{T}=\left(\begin{array}{c}
\vspace{0.2cm}
1/\sqrt{p_l}\sgn\left(\left(
\sum_{k=1}^{n}x_{k,j}\left(\bar{{\cal Y}}-\boldsymbol{y}_k\right)
\right)_i\right) \\
1/\sqrt{p_l}\sgn\left(\left(
\sum_{k=1}^{n}x_{k,j}\left(\bar{{\cal Y}}-\boldsymbol{y}_k\right)
\right)_i\right)\\
\vdots\\
1/\sqrt{p_l}\sgn\left(\left(
\sum_{k=1}^{n}x_{k,j}\left(\bar{{\cal Y}}-\boldsymbol{y}_k\right)
\right)_i\right) 
\end{array}\right)
$$
for any $i \in \{1,2,\cdots,m\}$, we get
\begin{equation*}
\begin{split}
\sup_{\begin{subarray}{c}
\|W_h;i:\|_2=1,\\
h=2,\cdots,l-1,\\
i=1,\cdots,p_{h+1};\\
h=l,i=1,\cdots,M.
\end{subarray}
}\left\| 
\nabla_{\boldsymbol{\theta}_1} {\cal L}_n({\cal Y}, \mu_{\boldsymbol{\theta}^{0}}(\boldsymbol{x}))\right\|_{\infty}
&=\max_{i,j}\left\{\sigma^{'}(0)^{l-1}\sum_{k=1}^{n}x_{k,j}\left(\bar{{\cal Y}}-\boldsymbol{y}_k\right)^{T}W_l \cdots W_{2;:i}\right\}\\
&=\sqrt{p_3\cdots p_l}\sigma^{'}(0)^{l-1}\max_{j} \sum_{t=1}^{m} \left|\left(
\sum_{k=1}^{n}x_{k,j}\left(\bar{{\cal Y}}-\boldsymbol{y}_k\right)
\right)_t\right|\\
&=\pi_l \sigma'(0)^{l-1} \|X^{\rm T} {\cal Y}_\bullet \|_\infty.
\end{split}
\end{equation*}

\section{Hessian Matrix} \label{app:Hessian}

Consider for simplicity the 3-layer network in regression with loss function
\begin{equation*}
{\cal L}_n({\cal Y}, \mu_{\boldsymbol{\theta}}({\cal X}))=\sqrt{
	\sum_{k=1}^{n}\left(y_k-(c+{\bf w}_{3}\sigma({\bf b}_{2}+W_{2}\sigma({\bf b}_{1}+W_1\boldsymbol{x}_k)))
	\right)^{2}}
	=: \sqrt{\sum_{k=1}^{n}G_{k}^{2}}.
\end{equation*}

Straightforward calculations lead to
\begin{equation*}
\begin{split}
\frac{\partial {\cal L}_n({\cal Y}, \mu_{\boldsymbol{\theta}}({\cal X}))}{\partial b_{1,i}}&=
\frac{1}{ {\cal L}_n({\cal Y}, \mu_{\boldsymbol{\theta}}({\cal X}))}\sum_{k=1}^{n}G_{k}\cdot\frac{\partial G_{k}}{\partial b_{1,i}},
\end{split}
\end{equation*}	
\begin{equation*}
\begin{split}
\frac{\partial^{2} {\cal L}_{n}({\cal Y}, \mu_{\boldsymbol{\theta}}({\cal X}))}{\partial b_{1,i}^{2}}&=
\frac{1}{-{\cal L}_{n}^{2}({\cal Y}, \mu_{\boldsymbol{\theta}}({\cal X}))}\cdot \frac{\partial {\cal L}_n({\cal Y}, \mu_{\boldsymbol{\theta}}({\cal X}))}{\partial b_{1,i}} \sum_{k=1}^{n}G_{k}\cdot\frac{\partial G_{k}}{\partial b_{1,i}}\\
&+\frac{1}{ {\cal L}_n({\cal Y}, \mu_{\boldsymbol{\theta}}({\cal X}))}\cdot \sum_{k=1}^{n}\left[\left(\frac{\partial G_{k}}{\partial b_{1,i}}\right)^{2}+G_{k}\cdot \frac{\partial^{2} G_{k}}{\partial b_{1,i}^{2}}\right],
\end{split}
\end{equation*}
and, for $i\neq j$,
\begin{equation*}
\begin{split}
\frac{\partial^{2} {\cal L}_{n}({\cal Y}, \mu_{\boldsymbol{\theta}}({\cal X}))}{\partial b_{1,i}\partial b_{1,j}}&=
\frac{1}{-{\cal L}_{n}^{2}({\cal Y}, \mu_{\boldsymbol{\theta}}({\cal X}))}\cdot \frac{\partial {\cal L}_n({\cal Y}, \mu_{\boldsymbol{\theta}}({\cal X}))}{\partial b_{1,j}} \sum_{k=1}^{n}G_{k}\cdot\frac{\partial G_{k}}{\partial b_{1,i}}\\
&+\frac{1}{ {\cal L}_n({\cal Y}, \mu_{\boldsymbol{\theta}}({\cal X}))}\cdot \sum_{k=1}^{n}\left[\frac{\partial G_{k}}{\partial b_{1,i}}\cdot \frac{\partial G_{k}}{\partial b_{1,j}}+G_{k}\cdot \frac{\partial^{2} G_{k}}{\partial b_{1,i}\partial b_{1,j}}\right].
\end{split}
\end{equation*}
Under condition ($L_0$), we get
\begin{equation*}
\begin{split}
\frac{\partial^{2} {\cal L}_{n}({\cal Y}, \mu_{\boldsymbol{\theta}}({\cal X}))}{\partial b_{1,i}^{2}}\bigg|_{L_0}&=\frac{n (\sigma^{'}(0))^{4}({\bf w}_3W_{2;:i})^2}{\|{\cal Y}_\bullet\|_2},
\end{split}
\end{equation*}
\begin{equation*}
\begin{split}
\frac{\partial^{2} {\cal L}_{n}({\cal Y}, \mu_{\boldsymbol{\theta}}({\cal X}))}{\partial b_{1,i}\partial b_{1,j}}\bigg|_{L_0}&=\frac{n (\sigma^{'}(0))^{4}({\bf w}_3W_{2;:i})({\bf w}_3W_{2;:j})}{\|{\cal Y}_\bullet\|_2}, 
\end{split}
\end{equation*}
\begin{equation*}
\begin{split}
\frac{\partial^{2} {\cal L}_{n}({\cal Y}, \mu_{\boldsymbol{\theta}}({\cal X}))}{\partial {\bf b}_{1}^2}\bigg|_{L_0}&=\frac{n (\sigma^{'}(0))^{4}({\bf w}_3W_{2})^{T}({\bf w}_3W_{2})}{\|{\cal Y}_\bullet\|_2}.
\end{split}
\end{equation*}
Consequently the Hessian with respect to ${\bf b}_1$ under condition ($L_0$) is positive semidefinite.
Similarly, one gets
\begin{equation*}
\begin{split}
\frac{\partial^{2} {\cal L}_{n}({\cal Y}, \mu_{\boldsymbol{\theta}}({\cal X}))}{\partial {\bf b}_{2}^2}\bigg|_{L_0}&=\frac{n (\sigma^{'}(0))^{2}{\bf w}_3^{T}{\bf w}_3}{\|{\cal Y}_\bullet\|_2},
\end{split}
\end{equation*}
showing that
the Hessian with respect to ${\bf b}_2$ at condition ($L_0$) is also positive semidefinite.

\section{Proof of Property~\ref{property:linearANN}}

Let ${\cal U}\subset {\mathbb R}$ be any subset of ${\mathbb R}$.
For all $u \in {\cal U}$, choosing $b=\max_{u \in {\cal U}} -u=-\min({\cal U})$, we have $u=\sigma_{\rm ReLU} (u+b)-b$.  Consider the data matrix $X$,  ${\bf u}=X{\boldsymbol\beta} \in {\mathbb R}^n$ and $b=-\min_{i=1,\ldots,n} u_i<\infty$.
Then for any point ${\bf x}\in {\mathbb R}^{p_1}$ in the convex hull of the $n$ data vectors $\{{\bf x}_i\}_{i=1,\ldots,n}$, we have $b+{\bf x}^{\rm T}{\boldsymbol \beta}=\sum_{i=1}^n \lambda_i (b+{\bf x}_i^{\rm T}{\boldsymbol\beta}) >0$,  
 so a linear function $\mu_{\boldsymbol \theta}^{\rm lin}({\bf x})=\beta_0+ {\bf x}^{\rm T}{\boldsymbol\beta}   = \beta_0 +\sigma_{\rm ReLU} ({\bf x}^{\rm T}{\boldsymbol\beta}+b)-b$ can be written as an ANN with a single neuron.

\bibliographystyle{plainnat}
\bibliography{article_bis}

\end{document}